\documentclass{article} 
\usepackage{iclr2023_conference,times}

\usepackage{amsmath,amsfonts,bm}

\def\eqref#1{equation~\ref{#1}}

\def\1{\bm{1}}

\DeclareMathAlphabet{\mathsfit}{\encodingdefault}{\sfdefault}{m}{sl}
\SetMathAlphabet{\mathsfit}{bold}{\encodingdefault}{\sfdefault}{bx}{n}

\usepackage{hyperref}
\usepackage{url}
\usepackage{booktabs}       
\usepackage{amsfonts}       
\usepackage{nicefrac}       
\usepackage{microtype}      
\usepackage{xcolor}         

\usepackage{amsmath}
\usepackage{multirow}
\usepackage{makecell}
\usepackage{colortbl}
\usepackage{graphicx}

\usepackage{subcaption}
\usepackage{alltt}
\usepackage{color}
\usepackage{algorithmicx,algorithm}

\title{Rethinking Knowledge Distillation via Cross-Entropy}

\author{Zhendong Yang\textsuperscript{$\star$}$^{1,2}$ \quad Zhe Li\textsuperscript{$\star$}\quad Yuan Gong$^{1}$\quad Tianke Zhang$^{1}$\quad\textbf{Shanshan Lao}$^{1}$\\
\textbf{Chun Yuan}\textsuperscript{$\dagger$}$^{1}$\quad\textbf{Yu Li}\textsuperscript{$\dagger$}$^{2}$\\
$^{1}$Tsinghua Shenzhen International Graduate School\\
$^{2}$International Digital Economy Academy  (IDEA)\\
{\tt\small \{yangzd21,gong-y21,ztk21,laoss21\}@mails.tsinghua.edu.cn}\\
{\tt\small axel.li@outlook.com \quad yuanc@sz.tsinghua.edu.cn \quad liyu@idea.edu.cn}
}

\iclrfinalcopy 
\begin{document}

\maketitle
\renewcommand{\thefootnote}{\fnsymbol{footnote}} 
\footnotetext[1]{Equal Contribution} 
\footnotetext[2]{Corresponding author} 

\begin{abstract}

Knowledge Distillation  (KD) has developed extensively and boosted various tasks. The classical KD method adds the KD loss to the original cross-entropy (CE) loss. We try to decompose the KD loss to explore its relation with the CE loss. Surprisingly, we find it can be regarded as a combination of the CE loss and an extra loss which has the identical form as the CE loss. However, we notice the extra loss forces the student's relative probability to learn the teacher's absolute probability. Moreover, the sum of the two probabilities is different, making it hard to optimize. To address this issue, we revise the formulation and propose a distributed loss. In addition, we utilize teachers' target output as the soft target, proposing the soft loss. Combining the soft loss and the distributed loss, we propose a new KD loss (NKD). Furthermore, we smooth students' target output to treat it as the soft target for training without teachers and propose a teacher-free new KD loss (tf-NKD). Our method achieves state-of-the-art performance on CIFAR-100 and ImageNet. For example, with ResNet-34 as the teacher, we boost the ImageNet Top-1 accuracy of ResNet18 from 69.90\% to 71.96\%. In training without teachers, MobileNet, ResNet-18 and SwinTransformer-Tiny achieve 70.04\%, 70.76\%, and 81.48\%, which are 0.83\%, 0.86\%, and 0.30\% higher than the baseline, respectively. The code is available at \url{https://github.com/yzd-v/cls_KD}.
\end{abstract}

\section{Introduction}
\label{intro}


Over the last decade, deep convolutional neural networks (CNNs) have significantly advanced the performance in many computer vision tasks~\citep{he2016deep,ren2015faster,ronneberger2015u,he2017mask}. Generally, a larger model scores higher but needs more computing resources. In contrast, a smaller model has less computation complexity and runs faster, but it performs less competitively than the larger one. Knowledge distillation  (KD) is proposed~\citep{hinton2015distilling} to bridge the gap to boost the small model in the training stage, causing no extra cost in the test time. Its core idea is when training the small student model; besides the supervision from the label, it also inherits the knowledge from the large teacher model as additional guidance. The distillation methods have been successfully applied to various domains, such as image classification~\citep{yang2020knowledge,zhou2020rethinking,chen2021distilling,zhao2022decoupled,yang2022masked}, object detection~\citep{chen2017learning,li2017mimicking,wang2019distilling,guo2021distilling,yang2021focal}, and semantic segmentation~\citep{liu2019structured,he2019knowledge,shu2021channel,yang2022masked}.

The classical distillation method~\citep{hinton2015distilling} utilizes the teacher's prediction as the soft label to guide the student. In addition to the predicted labels given by the teacher, there are also artificially given soft labels. Label smooth can be regarded as a particular case of the soft label and ~\citep{muller2019does} show it helps the models to represent the samples from the same class to the group in tight clusters. Tf-KD~\citep{yuan2020revisiting} explore the knowledge distillation from label smoothing regularization and propose a novel teacher-free knowledge distillation method.


However, previous works lack the consideration of the relation between the original CE loss and KD loss. From this perspective, we investigate the classical KD loss and find that the KD loss can be reformulated as a combination of the original CE loss and an extra loss. The extra loss mainly introduces the knowledge of all classes except the target class, which we call non-target distribution. Besides, the extra loss has the same form as CE loss. However, the extra loss aims to force the student's relative probability to be the same as the teacher's absolute probability. To this end, although the decomposition provides a term that looks like CE loss, the student's output should not be similar to the teacher's output when converging. To solve this problem, we modify the formula and proposed distributed loss to transfer the knowledge of non-target distribution.

Besides the non-target distribution, we believe the target information should also be reasonably introduced into the knowledge distillation. Inspired by the phenomenon that the soft label is easier than the ground-truth  (GT) label for a compact model to fit, we argue that the teacher's target output can also be viewed as the soft target. This soft target gives a much smoother label value than the GT label value. Based on the method, we apply the soft targets to the samples and propose the soft loss. 
Since the distributed loss provides non-target distribution knowledge and the soft loss provides the soft target knowledge, we can use these two losses in combination. In this way, we present our \textbf{N}ew \textbf{K}nowledge \textbf{D}istillation (NKD) loss. The students achieve significant improvements and state-of-the-art performance with our new loss.

Furthermore, as the student's prediction can also give the sample a much smoother label value, we try to use soft loss without teachers. Since student's predictions vary gradually during training, we smooth student's target output to make it more stable during training, propose our \textbf{t}eacher \textbf{f}ree \textbf{N}ew \textbf{K}nowledge \textbf{D}istillation  (tf-NKD) loss. 
We conduct various experiments to validate the effectiveness and robustness of our tf-NKD loss. Besides, the weights are all obtained by adjusting the student's target output. Therefore compared with the baseline, it does not take extra time to train a model.
\begin{figure}[t]
    \centering
    \includegraphics[width=\textwidth]{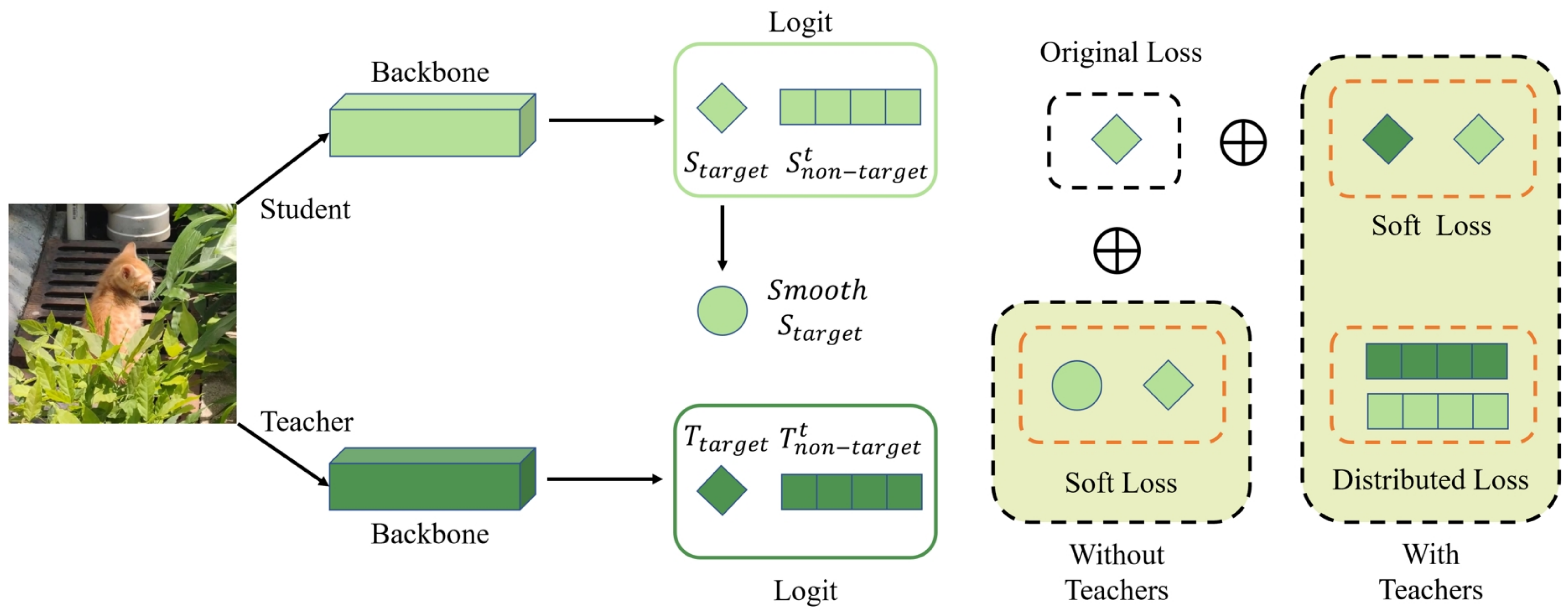}

    \caption{Illustration of the proposed NKD and tf-NKD.}
    \label{fig:structure}
    \vspace{-5mm}
\end{figure}

As we analyzed above, we propose a new paradigm of knowledge distillation loss, including distributed loss and soft loss. Combining these two losses with the original CE loss, we achieve state-of-the-art performance in various models. Besides, in the absence of teachers, we also get a considerable boost in many models using the soft loss. The two methods we propose are shown in Figure~\ref{fig:structure}. In a nutshell, the contributions are as follows:

\begin{itemize}
  \item
  We demonstrate that the classical KD loss can be regarded as the combination of the original CE loss and an extra loss. The extra loss mainly introduces the knowledge of non-target distribution and is hard to optimize. To address this issue, we propose distributed loss.
  \item
  Inspired by the soft label method, we propose soft loss, which uses the teacher's target output as the soft target for distillation. Combining distributed and soft loss, we propose NKD loss, achieving state-of-the-art performance.
  \item
  We smooth the student's target output as the soft target to train the students without teachers. In this way, we propose tf-NKD loss, which can also bring considerable improvements for the students without extra time costs.
\end{itemize}

\section{Related work}
\label{related work}

Knowledge distillation (KD) is a method to improve the model while keeping the network structure unchanged. It was first proposed by Hinton et al.~\citep{hinton2015distilling}, where the student is supervised by the hard labels and the soft labels from the teacher's output. Many following works focus on making better use of soft labels to transfer more knowledge. WSLD~\citep{zhou2020rethinking} analyzes soft labels and distributes different weights for them from a perspective of bias-variance trade-off. DKD~\citep{zhao2022decoupled} divides the classical KD according to the teacher's prediction and modifies the formulation of KD, achieving state-of-the-art performances. SRRL~\citep{yang2020knowledge} forces the output logits of teacher's and student's features after the teacher's linear layer to be the same.

Besides distillation on logits, some works aim at transferring knowledge from intermediate features. FitNet~\citep{romero2014fitnets} distills the semantic information from intermediate feature directly. OFD~\citep{heo2019comprehensive} designs the margin ReLU and modifies the measurement for the distance between students and teachers. RKD~\citep{park2019relational} extracts the relation from the feature map. CRD~\citep{tian2019contrastive} applies contrastive learning to distillation successfully. KR~\citep{chen2021distilling} transfers knowledge from multi-level features for distillation. SRRL~\citep{yang2020knowledge} utilizes the teacher's classifier to train the student's feature. MGD~\citep{yang2022masked} proposes a new distillation method that makes the student generate the teacher's feature instead of mimicking.


\section{Method}
\label{method}
\subsection{Distillation with teachers}
\label{kd with t}

Using $t$ denotes the target class, $C$ denotes the number of classes, $V_{i}$ denotes the label value, and $S_{i}$ denotes the student's output probability. The original loss for image classification can be formulated:
\begin{align}
    L_{ori} = -\sum_{i}^{C}V_{i}log (S_{i})=-V_{t}log ( S_{t})=-log (S_{t}).
\end{align}
Using $\lambda$ denotes the temperature for knowledge distillation, $T_{i}$ denotes the teacher's output probability. The classical KD loss can be formulated as:
\begin{align}
\label{eq:original kd begin}
    L_{kd}=-\sum_{i}^{C} T_{i}^{\lambda}log (S_{i}^{\lambda})&=-\sum_{i}^{C} T_{i}^{\lambda}log (S_{t}^{\lambda}\ast\frac{S_{i}^{\lambda}}{S_{t}^{\lambda}})\\
    &=-\sum_{i}^{C} T_{i}^{\lambda}log (S_{t}^{\lambda})-\sum_{i}^{C} T_{i}^{\lambda}log (\frac{S_{i}^{\lambda}}{S_{t}^{\lambda}}).
\end{align}
Because $\sum_{i}^{C} T_{i}^{\lambda}=\sum_{i}^{C} S_{i}^{\lambda}=1$ and $T_{t}^{\lambda}log (S_{t}^{\lambda}/S_{t}^{\lambda})=0$, $L_{kd}$ can be simplified as:
\begin{align}
\label{eq:original kd end}
    L_{kd}&=-log (S_{t}^{\lambda})-\sum_{i \ne t}^{C}T_{i}^{\lambda}log (\frac{S_{i}^{\lambda}}{S_{t}^{\lambda}}).
\end{align}
As $L_{kd}$ shows, $-log (S_{t}^{\lambda})$ has the same form as $L_{ori}$ and does not bring new knowledge for the student. While for the extra loss $-\sum_{i \ne t}^{C}T_{i}^{\lambda}log (S_{i}^{\lambda}/S_{t}^{\lambda})$, it has the same form as CE loss $-\sum p (x)log (q (x))$ and mainly introduces the non-target knowledge to the student. The CE loss aims at making $q (x)$ to be the same as $p (x)$. Therefore the sum of the two distributions needs to be equal. However, $T_{i}^{\lambda}$ is the absolute probability and $\sum_{i\ne t}^{C} T_{i}^{\lambda}= 1-T_{t}^{\lambda}$. While $S_{i}^{\lambda}/S_{t}^{\lambda}$ is the relative probability and $\sum_{i\ne t}^{C}S_{i}^{\lambda}/S_{t}^{\lambda}= (1-S_{t}^{\lambda})/S_{t}^{\lambda}$. So $S_{i}/S_{t}$ is hard to be similar to $T_{i}$.

To transfer the non-target knowledge and meet this constraint, we propose distributed loss:
\begin{align}
\label{eq:distributed loss}
    L_{distributed} = - \sum_{i\ne t}^{C}\hat T_{i}^{\lambda}log (\hat S_{i}^{\lambda}).
\end{align}
\begin{align*}
    \hat T_{i}^{\lambda} = \frac{T_{i}^{\lambda}}{1-T_{t}^{\lambda}},~~~\hat S_{i}^{\lambda} = \frac{S_{i}^{\lambda}}{1-S_{t}^{\lambda}}.
\end{align*}
In this case, we can see $\sum_{i\ne t}^{C}\hat T_{i}^{\lambda} = \sum_{i\ne t}^{C}\hat S_{i}^{\lambda}=1$, making the student learn teacher's non-target knowledge easier.

However, $L_{distributed}$ lacks the teacher's target knowledge. Some previous KD methods~\citep{hinton2015distilling, zhao2022decoupled} have proved that the teacher's predictions can be utilized as soft labels to accelerate the convergence and improve student performance. Inspired by this soft label method, we regard the teacher's target output probability $T_{t}$ as the soft target directly. Based on the soft target the teacher introduces, we propose the soft loss for distillation with teachers:
\begin{align}
\label{eq:soft loss}
    L_{soft} = -T_{t}log (S_{t}).
\end{align}

Finally, combining the original loss $L_{ori}$, distributed loss $L_{distributed}$ and soft loss $L_{soft}$, we propose our \textbf{N}ew \textbf{K}nowledge \textbf{D}istillation (NKD) loss as follows:
\begin{align}
\label{eq:new kd loss}
    L_{NKD} = -log (S_{t}) - T_{t}log (S_{t}) - \alpha\ast \lambda^{2}\ast\sum_{i\ne t}^{C}\hat T_{i}^{\lambda}log (\hat S_{i}^{\lambda}),
\end{align}
where $\alpha$ is a hyper-parameter to balance the loss.

\subsection{Training without teachers}
\label{kd without t}

The weight $T_{t}$ and $\hat T_{i}^{\lambda}$ in Equation~\ref{eq:new kd loss} are both got from teacher's output. Therefore $L_{NKD}$ is just suitable for distillation with teachers. The $\hat T_{i}^{\lambda}$ is about the non-target distribution knowledge, which needs to be calculated from a trained model. However, the soft target $T_{t}$ is the target output probability for the input image. We wonder if the soft target can be provided by adjusting the student's target output $S_{t}$. The difference is that $T_{t}$ is fixed and $S_{t}$ varies gradually during training. From this perspective, we adjust $S_{t}$, making it smoother to fit the training setting without teachers. This method can be applied to different models directly and does not take any extra time compared with the baseline. For the training without teachers, we propose our \textbf{t}eacher \textbf{f}ree \textbf{N}ew \textbf{K}nowledge \textbf{D}istillation (tf-NKD) loss:
\begin{align}
\label{eq:modified soft loss}
    L_{tf-NKD} = -log (S_{t}) -  (S_{t}+V_{t}-mean (S_{t}))log (S_{t}),
\end{align}
where $V_{t}$ denotes the target label value for the sample and $mean(\cdot)$ is calculated across different samples in a batch. Comparing with $L_{soft}$ in Equation~\ref{eq:soft loss}, we replace the $T_{t}$ with adjustable $S_{t}$. The experiments of $L_{tf-NKD}$ are shown in Subsection~\ref{soft loss}. We also discuss the effects of different ways to smooth the $S_{t}$ in Subsection~\ref{soft weight without t}. 

In short, we propose $L_{NKD}$ for distillation with teachers and $L_{tf-NKD}$ for training without teachers.


    
    
    
    

\section{Experiments}
\label{main experiments}

\subsection{Datasets and Details}
We conduct the experiments on CIFAR-100~\citep{krizhevsky2009learning} and ImageNet~\citep{deng2009imagenet}, which contains 100 and 1000 categories, respectively. For CIFAR-100, we use the 50k images for training and 10k for validation. For ImageNet, we use 1.2 million images for training and 50k images for validation. In this paper, we use accuracy to evaluate all the models.

For distillation with teachers, NKD has two hyper-parameters $\alpha$ and $\lambda$ in Equation\ref{eq:new kd loss}. For all the experiments, we adopt $\{\alpha = 1.5, \lambda=1\}$ on ImageNet. The training setting for distillation is the same as training the students without distillation. We use 8 Tesla-V100 GPUs to conduct the experiments with MMClassition~\citep{2020mmclassification} based on Pytorch~\citep{paszke2019pytorch}. While for CIFAR-100, we follow the training setting from DKD~\citep{zhao2022decoupled}.

\subsection{Distillation with teachers}

\begin{table*}[t]
\centering
\caption{Results of different distillation methods on the CIFAR-100 validation. The models in the first row and the second row are the teacher and student, respectively.}
\begin{tabular}{c|c|cccccc|c}
\toprule
Method&Type & \makecell{VGG13\\VGG8} & \makecell{ResNet32x4\\ResNet8x4}&\makecell{VGG13\\MobileNetV2}&\makecell{ResNet50\\MobileNetV2}&\makecell{ResNet32x4\\ShuffleNetV1}\\
\midrule
baseline&-&70.36&72.50&64.60&64.60&70.50\\
\midrule
RKD& Feature&71.48&71.90&64.52&64.43&72.28\\
CRD& Feature&73.94&75.51&69.73&69.11&75.11\\
OFD& Feature&73.95&74.95&69.48&69.04&75.98\\
KR& Feature&74.84&75.63&{\bf70.37}&69.89&{\bf77.45}\\
KD& Logit&72.98&73.33&67.37&67.35&74.07\\
WSLD& Logit&74.36&76.05&69.02&70.15&75.46\\
DKD& Logit&74.68&76.32&69.71&70.35&76.45\\
Ours& Logit&{\bf74.86}&{\bf76.35}&70.22&{\bf70.67}&76.54\\
\bottomrule
\end{tabular}
\label{table:cifar results}
\end{table*}

For the distillation with teachers, we first conduct experiments with various teacher-student distillation pairs on CIFAR-100, shown in Table~\ref{table:cifar results}. In this setting, we evaluate our method on various models with different architectures including VGG~\citep{simonyan2014very}, ResNet~\citep{he2016deep}, MobileNetV2~\citep{sandler2018mobilenetv2} and ShuffleNet~\citep{zhang2018shufflenet}. We compare our method with the classical KD~\citep{hinton2015distilling} and several other state-of-the-art distillation methods for both heterogeneous and homogeneous distillation. As the results show, our method brings the students remarkable accuracy gains over other methods. For both heterogeneous and homogeneous distillation, our method achieves the best performance among logit-based distillation and even surpasses feature-based distillation in some distillation settings.


\begin{table*}[t]
  \setlength{\tabcolsep}{8.5 pt}
  \centering
  \caption{Results of different distillation methods on ImageNet dataset. {\textbf{T}} and {\textbf{S}} indicate the teacher and student, respectively.}
  \begin{tabular}{c|l|cc|l|cc}
    \toprule
    Type& Method & Top-1  & Top-5 & Method &Top-1&Top-5\\
    \midrule
    &ResNet-34 (\textbf{T}) & 73.62 &91.59&ResNet-50 (\textbf{T})&76.55&93.06\\
    &ResNet-18 (\textbf{S}) & 69.90 &89.43&MobileNet (\textbf{S})&69.21&89.02\\
    \midrule
    \multirow{6}{*}{\makecell{Feature}}
    &AT & 70.59&89.73&AT&70.72&90.03\\
    &OFD & 71.08&90.07&OFD&71.25&90.34\\
    &RKD & 71.34&90.37&RKD&71.32&90.62\\
    &CRD & 71.17&90.13&CRD&71.40&90.42\\
    &KR & 71.61&90.51&KR&72.56&91.00\\
    &MGD & 71.69&90.49&MGD&72.49&90.94\\
    \midrule
    \multirow{4}{*}{\makecell{Logit}}
    &KD & 71.03&90.05&KD&70.68&90.30\\
    &WSLD& 71.73&{\bf90.53}&WSLD&72.02&90.70\\
    &DKD & 71.70&90.41&DKD&72.05&{\bf91.05}\\
    &\cellcolor{lightgray!45}{\bf Ours} & \cellcolor{lightgray!45}{\bf 71.96}&\cellcolor{lightgray!45}90.48&\cellcolor{lightgray!45}{\bf Ours}&\cellcolor{lightgray!45}{\bf72.58}&\cellcolor{lightgray!45}90.96\\
    \midrule
    \multirow{2}{*}{\makecell{Feature $+$ Logit}}
    &SRRL & 71.73&90.60&SRRL&72.49&90.92\\
    &\cellcolor{lightgray!45}{\bf Ours}+MGD & \cellcolor{lightgray!45}{\bf 72.01}&\cellcolor{lightgray!45}{\bf 90.84}&\cellcolor{lightgray!45}{\bf Ours}+MGD&\cellcolor{lightgray!45}{\bf 73.10}&\cellcolor{lightgray!45}{\bf 91.32}\\
    \bottomrule
   \end{tabular}
  \label{table:imagenet results}
\end{table*}

To further demonstrate the effectiveness and robustness of our method, we also test it on a more challenging dataset, ImageNet. We set two popular teacher-student pairs, which include homogeneous and heterogeneous distillation. The homogeneous distillation is ResNet34-ResNet18, and the heterogeneous distillation is ResNet50-MobileNet. 

The results of different methods on ImageNet are shown in Table~\ref{table:imagenet results}. As the results show, our method outperforms all the previous methods by just distilling on the logit. Our method brings  consistent and significant improvements to the students for both distillation settings. The student ResNet18 and MobileNet achieve 71.96\% and 72.58\% Top-1 accuracy, getting 2.06\% and 2.37\% accuracy gains with the knowledge transferred from the teacher's logit, respectively. Furthermore, we try to combine our method with the SOTA feature-based distillation method MGD~\citep{yang2022masked} to explore the upper bound for the distillation pairs. In this way, the student ResNet18 and MobileNet can achieve 72.01\% and 73.10\% Top-1 accuracy, getting another 0.05\% and 0.52\% accuracy gains, respectively.

\begin{table*}[t]
  \centering
\setlength{\tabcolsep}{21 pt}
  \caption{Results of different smooth methods. {\bf Extra time cost} means whether needs a teacher, which needs the time for training a teacher first and inference the teacher during training.}
  \begin{tabular}{c|c|cc}
    \toprule
    Methods &Extra time cost & ResNet18 & ResNet50\\
    \midrule
    baseline       &-&69.90 &76.55\\
    \midrule
    Label smooth   &-&69.92 (+0.02) &76.64 (+0.09)\\
    Tf-KD          &-&70.14 (+0.24)&76.59 (+0.04)\\
    Our tf-NKD     &-&{\bf70.76 (+0.86)}&{\bf76.93 (+0.38)}\\
    \bottomrule
  \end{tabular}
  \label{table:tf-nkd compare}
\end{table*}

\begin{table*}[ht]
  \centering
\setlength{\tabcolsep}{16 pt}
  \caption{Top-1 accuracy of different models with our tf-NKD on ImageNet dataset.}
  \begin{tabular}{l|cc|cc}
    \toprule
    Model& Params (M)&Flops (G) & Baseline & + tf-NKD\\
    \midrule
    MobileNet    &4.2 &0.575 & 69.21 &70.04{ (+0.83)}\\
    MobileNetV2  &3.5 &0.319 & 71.86 &72.08{ (+0.22)}\\
    ShuffleNetV2 &2.3 &0.149 & 69.55 &69.93{ (+0.38)}\\
    \midrule
    ResNet-18    &11.69 &1.82 & 69.90 &70.76{ (+0.86)}\\
    ResNet-50    &25.56 &4.12 & 76.55 &76.93{ (+0.38)}\\
    ResNet-101   &44.55 &7.85 & 77.97 &78.30{ (+0.33)}\\
    \midrule
    Swin-Tiny    &28.29 &4.36 & 81.18 &81.48{ (+0.30)}\\
    Swin-Small   &49.61 &8.52 & 83.02 &83.08{ (+0.06)}\\
    Swin-Base    &87.77 &15.14 & 83.36 & 83.36\\
    \bottomrule
  \end{tabular}
  \vspace{2pt} 
  \label{table:sd results}
\end{table*}

\subsection{Training without teachers}
\label{soft loss}
Our tf-NKD is designed for training the students without teachers. The soft targets in tf-NKD are all obtained by smoothing the student's target output, so there is not any extra time cost compared with training the student directly. To evaluate the effectiveness of tf-NKD, we first compare it with two other methods that smooth the labels, including label smooth and Tf-KD~\citep{yuan2020revisiting}. As the results shown in Table\ref{table:tf-nkd compare}, tf-NKD brings a much larger improvement than the two methods. The label smooth and Tf-KD mainly introduce the knowledge about non-target distribution, while our tf-NKD is inspired by the KD method and brings the soft target knowledge.

To further evaluate the effectiveness and generalization of tf-NKD, we apply tf-NKD to various models with different architectures and sizes, including MobileNet~\citep{howard2017mobilenets}, MobileNetV2~\citep{sandler2018mobilenetv2}, ShuffleNetV2~\citep{ma2018shufflenet}, ResNet~\citep{he2016deep} and SwinTransformer~\citep{liu2021swin}. As the results are shown in Table~\ref{table:sd results}, the tf-NKD is beneficial to all different architectures, including lightweight models, CNN-based models and hybrid models. All the architectures can achieve Top-1 accuracy gains. Even for SwinTransformer-Tiny, it also brings 0.3\% accuracy gains. Besides, tf-NKD improves even more for lightweight models. For example, it brings MobileNet and ResNet18 0.83\% and 0.86\% Top-1 accuracy gains, while the performance of Swin-Base keeps the same. The results show our tf-NKD is general and effective.

\section{Analysis}
\subsection{Effects of soft loss and distributed loss for NKD}
In this paper, we propose soft loss and distributed loss, which transfer the knowledge of the teacher's soft target and non-target distribution, respectively. In this subsection, we conduct experiments on soft loss and distributed loss to investigate their influences. As shown in Table~\ref{table:ablation study}, the soft loss and distributed loss lead to 1.16\% and 1.48\% accuracy improvements respectively, which shows both the knowledge of the soft target and non-target distribution is helpful to the student. Besides, the knowledge that soft loss and distributed loss transfer are independent. Combining them together allows us to make better use of the teacher's knowledge. In this way, the student achieves 71.96\% Top-1 accuracy, which is significantly greater than using just using soft loss or distributed loss. 

Moreover, we modify classical KD and propose distributed loss for distillation. Here we also compare the distributed loss with classical KD. As it shows, distributed loss brings 1.48\% gains while classical KD brings 1.03\%. The comparison also validates the effectiveness of distributed loss.
\begin{table}
  \centering
    \setlength{\tabcolsep}{18 pt}
  \caption{Ablation study of distributed loss and soft loss.}
  \begin{tabular}{c|c|c|ccc}
    \toprule
    Method & \multicolumn{4}{c}{ResNet34 - ResNet18}\\
    \midrule
    soft~~loss     & - & - &\checkmark&-&\checkmark\\
    distributed~~loss  & - & - &-&\checkmark&\checkmark\\
    classical~~KD      &-& \checkmark &-&-&-\\
    \midrule
    Top-1~~Acc & 69.90 &71.03 & 71.06 & 71.38 & {\bf71.96}\\
    Top-5~~Acc & 89.43 &90.05 & 89.51 & 90.47 & {\bf90.48}\\
    \bottomrule
  \end{tabular}
  \label{table:ablation study}
\end{table}

\subsection{Soft loss with different soft target}
\label{soft weight with t}

For the soft loss, we utilize the teacher's target output $T_{t}$ as the soft target in Equation~\ref{eq:soft loss}. In this subsection, we explore the effects of different soft targets $T_{t}$ for the soft loss.

We conduct experiments by distilling ResNet50 on ImageNet, which is shown in Table~\ref{table:different teachers}. The performance of the teachers varies from MobileNetV2's 71.86\% to ResNet50's 76.64\%. We also design a perfect teacher, which has 100\% accuracy and the $T_{t}$ for each sample is 1. For the teacher trained without label smooth, a better teacher's label output probability is closer to gt-label. For example, the same sample's $T_{t}$ of ResNet-50 is closer to 1 than that of ResNet-34. That is to say: strong teachers have high target output values, making the `soft target' harder. Such harder `soft target' bring the student fewer improvements. As the results show, the worst teacher MobileNetV2 brings 0.43\% gains while ResNet50 just brings 0.33\%. When we use the perfect teacher, which means $T_{t} = 1$, the soft loss even harms the student and causes a 0.1\% Top-1 accuracy drop.

However, when the teacher is trained with label smooth, the teacher with high performance can also benefit the student a lot. Specifically, the accuracy of ResNet-50 with label smooth is 76.64\%, which is 0.09\% higher than ResNet-50's 76.55\%. But it brings the student 0.54\% Top-1 accuracy gains, while ResNet-50 just brings 0.33\% Top-1 accuracy gains. We try to decompose the cross-entropy loss with label smooth to explore this phenomenon:
\begin{align}
  L_{ls} &= -\sum_{i}^{C}V_{i}log ( S_{i}) \\
     &= -\sum_{i}^{C}V_{i}log (\frac{S_{i}}{S_{t}})- \sum_{i}^{C}V_{i}log (S_{t}).
\end{align}
Because $\sum_{i}^{C} V_{i}=1$, $V_{i} = \alpha/C (i\ne t)$ and $V_{t}log (S_{t}/S_{t})=0$, $L_{ls}$ can be simplified as:
\begin{align}
\label{eq:ls}
  L_{ls} = -\frac{\alpha}{C}\sum_{i\ne t}^{C} log (\frac{S_{i}}{S_{t}})-log (S_{t}),  
\end{align}
where $C$ is the number of classes and $\alpha$ is a hyper-parameter. As the equation shows, the loss with label smooth can be regarded as the combination of the original cross-entropy loss and extra loss on the non-target output. The extra loss can prevent the model from over-fitting, which will decrease $T_{t}$ and get a smoother output. Such models can get softer `soft targets' and thus benefit the student more through the soft loss.

\begin{table*}[t]
  \centering
  \setlength{\tabcolsep}{7.5 pt}
  \caption{Results of distilling ResNet-50 only by soft loss with different teachers on ImageNet dataset. The perfect teacher means the model has 100\% accuracy and the target prediction $T_{t}$ for each image is 1. {\bf ls} means training the model with label smooth. {\bf T} and {\bf S} means the teacher and student.}
  \begin{tabular}{l|c|c|cc}
    \toprule
    Type& Extra time cost&Top-1 Acc (\textbf{T}) & Top-1 Acc (\textbf{S}) & Top-5 Acc (\textbf{S})\\
    \midrule
    baseline       &-& -    &76.55 &93.06\\
    \midrule
    $S_{t}+V_{t}-mean (S_{t})$  &-& -    &76.93 (+0.38) &93.21\\
    \midrule
    MobileNetV2    &$\checkmark$&71.86 &76.98 (+0.43) &93.22\\
    ResNet-34      &$\checkmark$&73.62 &76.89 (+0.34) &93.22\\
    ResNet-50      &$\checkmark$&76.55 &76.88 (+0.33) &93.16\\
    ResNet-50 (ls)  &$\checkmark$&76.64 &77.09 (+0.54) &93.26\\
    Perfect T      &-&100.0 &76.45 (-0.10) &93.06\\
    \bottomrule
  \end{tabular}
  \label{table:different teachers}
\end{table*}

\subsection{Different ways to smooth student's target output for tf-NKD}
\label{soft weight without t}
For tf-NKD, we replace the weights of NKD's soft loss by smoothing the student's target output $S_{t}$ according to Equation~\ref{eq:modified soft loss}. In this subsection, we explore the effects of different ways to smooth the student's target output $S_{t}$.

\begin{table*}[t]
  \centering
    \setlength{\tabcolsep}{18 pt}
  \caption{Results of training ResNet18 directly with tf-NKD on ImageNet dataset. The weights for tf-NKD are obtained by smoothing the student's target prediction $S_{t}$ in different ways. $V_{t}$ means the gt-label value for the input sample. {\bf mean}, {\bf sum}, {\bf max} and {\bf softmax} are all calculated with different samples in a training batch.}
  \begin{tabular}{l|c|cc}
    \toprule
    smooth ways &Extra time cost & Top-1 Acc & Top-5 Acc\\
    \midrule
    baseline       &-&69.90 &89.43\\
    \midrule
    $S_{t}$                   &-&70.50 (+0.60) &89.38\\
    $S_{t}+V_{t}-mean (S_{t})$     &-&{\bf 70.76 (+0.86)} &89.30\\
    $softmax (S_{t})*sum (v)$   &-&70.57 (+0.67) &89.34\\
    $\sqrt{S_{t}-min (S_{t})}$ &-&70.57 (+0.67) &89.13\\
    $S_{t}/max (S_{t})$        &-&70.53 (+0.63) &89.39\\
    $S_{t}/mean (S_{t})$       &-&70.50 (+0.60) &88.99\\
    \midrule
    ResNet18      &$\checkmark$&{\bf 70.75 (+0.85)} &89.53\\
    \bottomrule
  \end{tabular}
  \vspace{2pt} 
  \label{table:different ways}
\end{table*}

We need a fixed label for training, such as 1 for an image or  (0.8,0.2) for a mixed-up image. However, the student's target output $S_{t}$ gradually increases during training. Especially, as shown in Fig 2, for easy samples, the $S_{t}$ is less than 0.2 at the beginning but larger than 0.9 at the last epochs. When the $S_{t}$ is small, the student's output can not describe the sample's true category actually. In this case, the purpose of optimizing the model is to learn a very small label. We propose two key points for smoothing the student's target output $S_{t}$: {\bf1) reflect the sample's true category in every epoch} and {\bf2) prevent the soft target of every sample varying greatly in different epochs}. 

Based on the two guidelines, we devise several different smooth methods. We conduct experiments by training ResNet18 on ImageNet to explore these smooth methods, which are shown in Table~\ref{table:different ways}. As the results show, tf-NKD with all the methods can bring the student considerable improvements. Especially, the model gets 0.86\% gains when using $S_{t}+V_{t}-mean (S_{t})$ as the weights, which is even higher than using the teacher ResNet18's output as the soft target. We finally apply this way to smooth the student's target output for tf-NKD, which is shown in Equation~\ref{eq:modified soft loss}. The results of applying tf-NKD to more models can be seen in Table~\ref{table:sd results}.

\begin{figure}[t]
    \centering
    \begin{subfigure}[b]{0.49\textwidth}
         \centering
         \includegraphics[width=\textwidth]{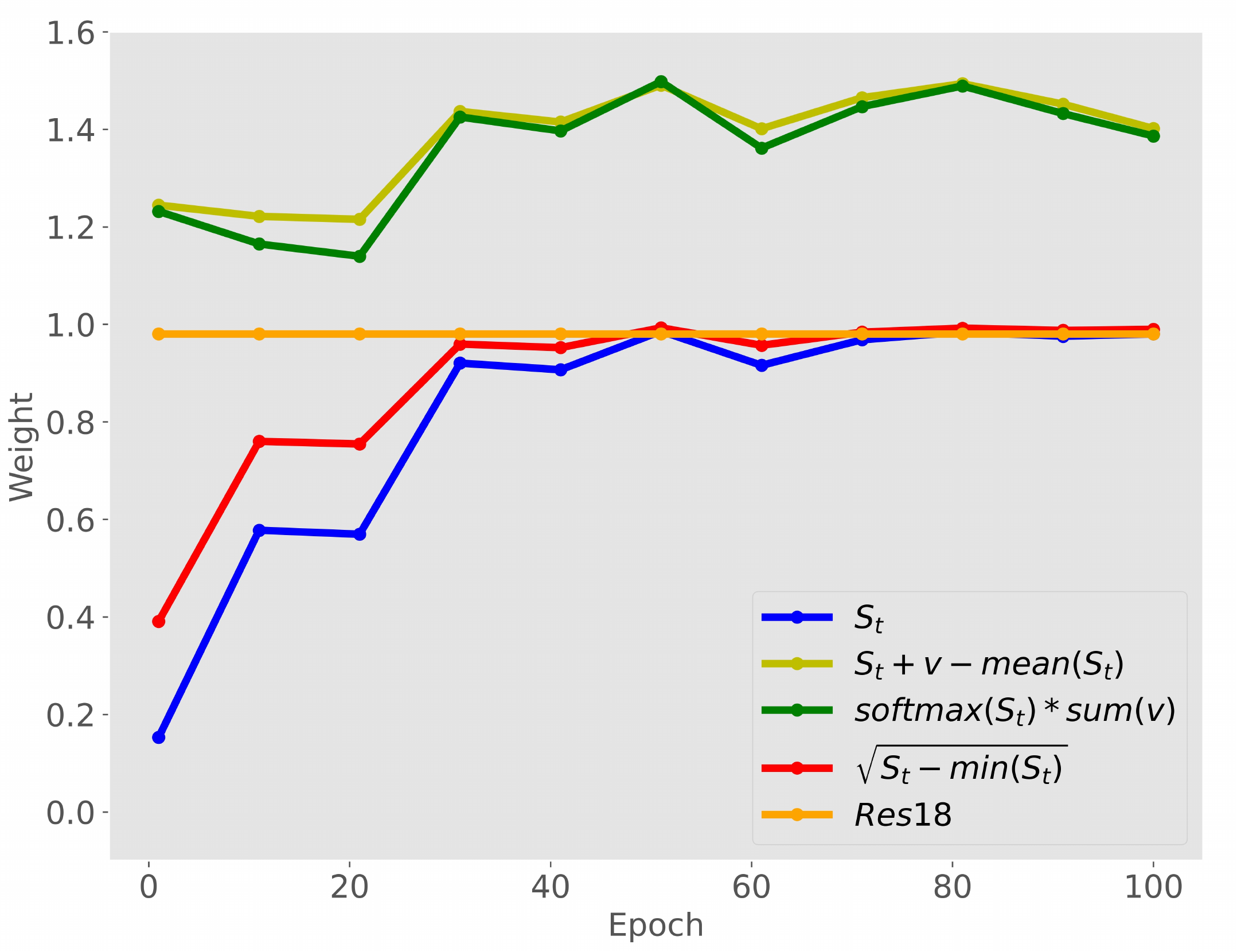}
         \caption{easy sample}
         \label{easy img}
     \end{subfigure}
     \begin{subfigure}[b]{0.49\textwidth}
         \centering
         \includegraphics[width=\textwidth]{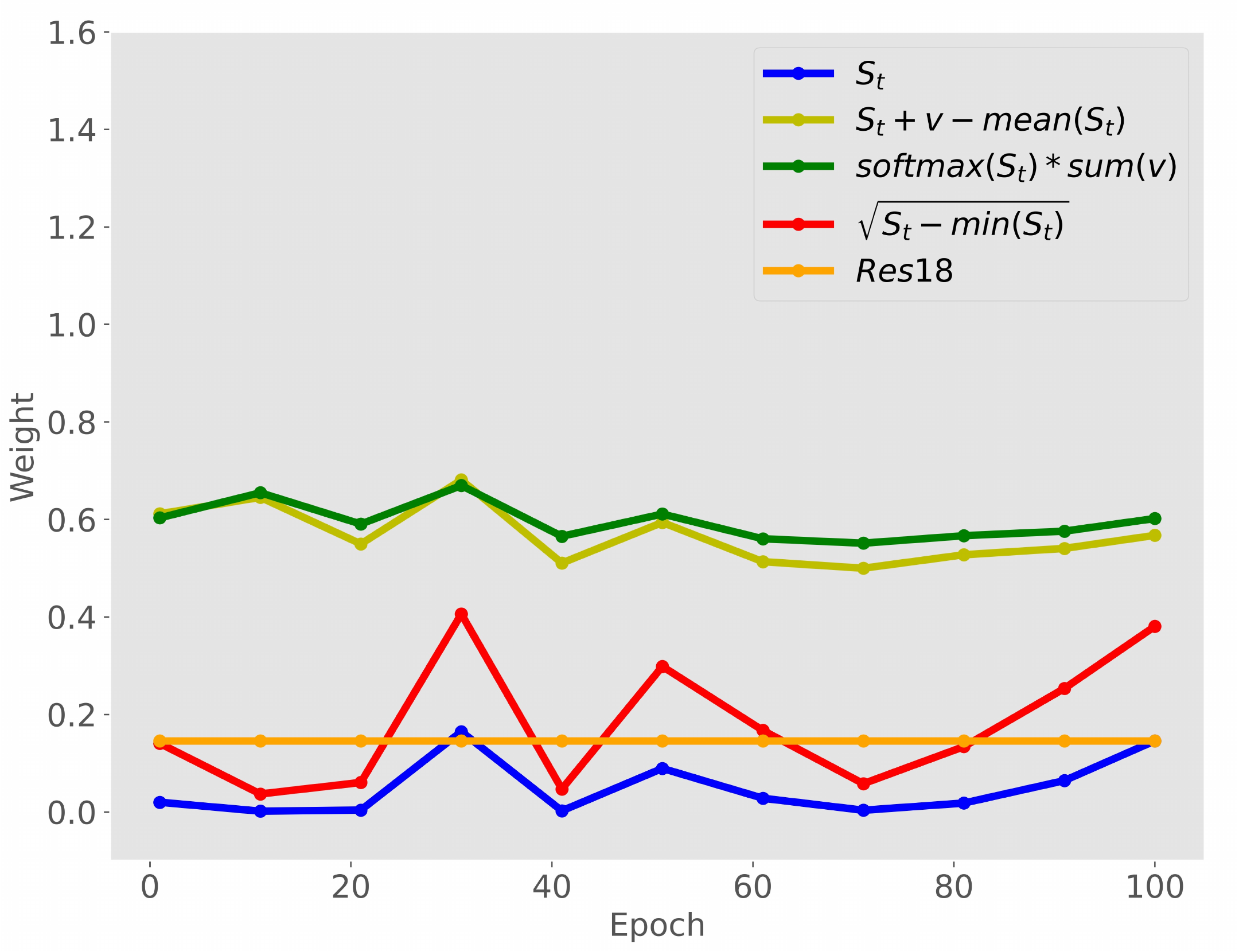}
         \caption{hard sample}
         \label{hard img}
     \end{subfigure}
    \caption{The curve of the weights of two representative samples during training. Different formulations mean adjusting the $S_{t}$ in different ways.}
    \label{fig:sample visualization}
\end{figure}

Furthermore, we select two representative samples and visualize the soft targets calculated in different ways in different epochs. As Figure~\ref{fig:sample visualization} shows, the student's target output $S_{t}$ is very small even for the easy samples at the beginning and varies greatly from different epochs. This makes it hard to represent different samples' true categories accurately. However, as Figure~\ref{fig:sample visualization} shows, the weights obtained by smoothing $S_{t}$ according to Equation~\ref{eq:modified soft loss} {\bf1) reflect the true category of different samples in every epoch} and {\bf2) vary smoothly in different epochs}. In this way, we can utilize the weights to get a better model with the same training time as the baseline. 

\begin{table*}[t]
\centering
\setlength{\tabcolsep}{17 pt}
\caption{Results on the CIFAR-100 validation with different temperature. We use ResNet-34 as the teacher to distill the student ResNet-18.}
\begin{tabular}{c|cccccc}
\toprule
$\lambda$ & 0.5 & 1.0 & 2.0 & 3.0 & 4.0 & 5.0\\
\midrule
Top-1 & 80.42 & 80.55 & {\bf 80.76} & 80.72 & 80.54 & 80.50\\
Top-5 & 94.89 & 95.14 & {\bf 95.14} & 95.11 & 95.05 & 95.11\\
\bottomrule
\end{tabular}
\label{table:ablation T}
\end{table*}

\subsection{The effect of the temperature}
\label{sec:temperature}
The temperature $\lambda$ in Equation~\ref{eq:new kd loss} is a hyper-parameter used to adjust the distribution of the teacher's logit. KD always applies $\lambda>1$ to make the logit become more smooth, which causes the logit contains more non-target distribution knowledge. The target output probability of the same model will get a higher value on an easy dataset, such as CIFAR-100. This causes $\hat T_{i}^{\lambda}$ in Equation~\ref{eq:new kd loss} contains less knowledge, which may bring adverse effect to the distillation. In this subsection, we explore the effects by using different temperatures to distill the student ResNet18 on CIFAR-100, which is shown in Table~\ref{table:ablation T}. The results show that temperature is an important hyper-parameter.

\section{Conclusion}
\label{sec:con}

In this paper, we first analyze the relation between classical KD loss and original CE loss. From this point of view, we modify the formulation of KD and propose distributed loss to transfer the knowledge of non-target distribution. Besides, we propose soft loss, which regards the teacher's target output as the soft target for the student to learn. We propose New Knowledge Loss (NKD) which includes distributed and soft loss, helping students achieve state-of-the-art performance. Furthermore, we smooth student's target output as the soft target to train the model directly, which also brings the students considerable improvements without teachers or extra time costs. 

\subsubsection*{Limitations}
We try several attempts to smooth the student's target output for tf-NKD. However, the way to adjust the student's target output to the sample's soft target is still naive and left as future work. Moreover, we transfer the target knowledge from the student to the student successfully. We can consider whether it is available to transfer the non-target knowledge from the student to the student too. We believe it's meaningful work and worth exploring.



\bibliography{iclr2023_conference}
\bibliographystyle{iclr2023_conference}


\end{document}